\pdfoutput=1
\documentclass{llncs}

\usepackage{refcount}
\usepackage{times}
\usepackage{url}
\usepackage{latexsym}
\usepackage{comment}
\usepackage{multirow}
\usepackage[utf8]{inputenc}

\usepackage{amsmath}
\usepackage{graphicx}
\usepackage{url}

\title{WS4A: a Biomedical Question and Answering System based on public Web Services and Ontologies}

\author{Miguel J. Rodrigues \thanks{Corresponding author: mrodrigues@lasige.di.fc.ul.pt} \and Miguel Falé \and  Andre Lamurias \and Francisco M. Couto 
}

\authorrunning{Rodrigues et al.} % abbreviated author list (for running head)
\institute{LaSIGE, Faculdade de Ci\^encias, Universidade de Lisboa, Portugal}

\begin{document}
\maketitle

\begin{abstract}
This paper describes our system, dubbed WS4A (Web Services for All),
that participated in the fourth edition of the BioASQ challenge (2016). 
We used WS4A to perform the Question and Answering (QA) task 4b, which consisted on the retrieval of relevant concepts, documents, snippets, RDF triples, exact answers and "ideal answers" for each given question. The novelty in our approach consists on the maximum exploitation of existing web services in each step of WS4A, such as the annotation of text, and the retrieval of metadata for each annotation. The information retrieved included concept identifiers, ontologies, ancestors, and most importantly, PubMed identifiers. The paper describes the WS4A pipeline and also presents the precision, recall and f-measure values obtained in task 4b. Our system achieved two second places in two subtasks on one of the five batches.
\end{abstract}


\section{Introduction}

This paper describes our participation in the BioASQ challenge edition of 2016\footnote{\url{http://www.bioasq.org/participate/challenges}}, 
a challenge on large-scale biomedical semantic indexing and question answering. Our participation focused on the Question and Answering (QA) task 4b, which consisted on the retrieval of relevant concepts, documents, snippets, RDF triples, exact answers and "ideal answers" for each given question. The competition was composed of two phases: in the first one the goal was to take as input a set of queries, and respond with set of the most relevant concepts, articles, snippets and RDF triples. 
The ontologies and terminologies from where the list of concepts were to be retrieved are: the Medical Subject Headings (MeSH); the Gene Ontology (GO); the Universal Protein Resource (UniProt); the Joint Chemical Dictionary (Jochem); the Disease Ontology (DO).
%(from Linked Life Data project\footnote{\url{http://linkedlifedata.com}}), all in English; 
In the second phase, gold responses to the questions answered of the first phase were provided, so this time, the goal was to provide a response by exploring the correct documents and snippets gathered by these experts. This challenge required the use of selected ontologies and answers in a required format according to the type of query.
%tirar este ----- There were also the following types of questions: Yes/no questions, that require only "yes" or "no" answers; Factoid questions, that require some kind of information or expression; List questions, could be a list of factoids; Summary questions, that require a short text as an answer.

We developed the system WS4A (Web Services for (4) All), which novelty was to explore every possible option to use public web services and incorporate available domain knowledge. WS4A addressed the above tasks by first recognizing relevant terms in the query and also in the abstracts associated with it based on available Web Services.
Next, the system mapped those terms to the respective concepts in ontologies and terminologies presented below. Then, WS4A compared those concepts to identify the responses that shared most concepts with the ones associated to the query. WS4A employed semantic similarity to measure how close in meaning they are even if they do no share the same exact concepts. Additionally, WS4A used Machine Learning \cite{pedregosa2011scikit} techniques to classify if an abstract is either relevant or not for the given query.

%In Section 1 we introduce the topic and the challenge at hand.
Section~\ref{webservices} describes the exploited Web Services by WS4A, and Section~\ref{pipeline} explains its composing modules. In Sections 4 and 5, we present and discuss our results, along with future work. 
%\section{Related Work} \label{relatedwork}
%This section does intend to be a comprehensive review of all previous works in this area,  but an overview of other recent and related %systems. In CLEF 2015 \cite{clef2015} among the systems that entered the BioASQ task 3b \cite{balikas2015results}, we noted that OAQA %\cite{OAQA2015} used supervised learning techniques to predict answer types (answer type coercion) and score candidate answers.

%This section describes the Web Services that WS4A used to annotate the questions, abstracts and PubMedIds related to those annotations, and explain the retrieved data transitions throughout the WS4A pipeline.
% We searched for Web Services of the retrieved annotations and found that the ChEBI and PUBC where the ease of implementation due to documentation found.
%\subsubsection{BioPortal}

\section{Web Services} \label{webservices} 
The first Web Service used was the one provided by BioPortal \cite{noy2009bioportal}, that given a text returns the ontology concepts mentioned in it. There are some other parameters that WS4A explored, but with the purpose of better filtering the results, such as: i) longest annotation only (set to false); ii) number exclusion (set to false); iii) whole word only (set to true) and iv) synonym exclusion (set to false). The results were provides as a JSON dictionary, divided by annotation and ontology. 

% All the examples given were strictly used in the development of this study, and there are no more other Web Services used besides the ones presented in this section.

%\url{"http://data.bioontology.org/annotator?apikey=586083bb-722f-4c1c-90d2-799af9629f2e&text=What%20is%20the%20association%20of%20estrogen%20replacement%20therapy%20and%20intracranial%20meningioma%20risk%3F&ontologies=GO,DOID,MESH&longest_only=false&exclude_numbers=false&whole_word_only=true&exclude_synonyms=false"}
%\subsubsection{Uniprot}
%\footnote{http://www.uniprot.org/uniprot/}

With the UniProt Web Services \cite{uniprot2014uniprot}, we may gather PubMedIds from protein descriptions, but to get these, we need to use another Web Service in order the obtain this description, the  Whatizit\footnote{http://www.ebi.ac.uk/webservices/whatizit/info.jsf} Web Service.
%\subsubsection{Whatizit} 

Whatizit \cite{rebholz2008text} is an alternative to BioPortal, when it comes to identifying specific terms in sentences. The Web Service present in Whatizit identifies the words in a sentence according to a given vocabulary and as a result provides a XML with an identifier of the Uniprot database in one of its tags. 

Back to UniProt, the request is made in the following format: i) the Web Service URL; ii) followed by a protein identifier; iii) ending with the available format that we chose (from our understanding, XML and HTML formats were available). For example: \url{http://www.uniprot.org/uniprot/P12345.xml}
%\subsubsection{PubChem}

When it comes to retrieving PubMed identifiers (PubMed IDs), the PubChem Web Service \cite{wang2009pubchem} was the last one to be used. This Web Service provided an easy interface, since it required trivial parameters and URLs. Each request was structured in the following way: 
\begin{itemize}
\item Base URL: \url{http://pubchem.ncbi.nlm.nih.gov/rest/pug/compound/name/}
\item A word in the question
\item Data Format: \url{xrefs/PubMedID/JSON}. Depending on the word, the result set (in XML) would be empty or not. If not, it meant that some PubMed IDs were present.
\end{itemize}
An example of the request for "oxygen" in the JSON data format would be: \url{https://pubchem.ncbi.nlm.nih.gov/rest/pug/compound/name/oxygen/xrefs/PubMedID/JSON},

%\subsection{NCBI}
%\footnote{http://eutils.ncbi.nlm.nih.gov}

NCBI has a service, eutils \cite{sayers2004building}, that provides a broad range of options. Two of them were used in order to retrieve all PubMed articles by their identifiers. The dates of these articles had to be before November 19th, 2015. The URLs are easy to identify and build, since they have a specific format, for example \url{http://www.ncbi.nlm.nih.gov/pubmed/23687640} where 223687640 is the PubMed ID. From eutils we used two services, one that searches for PubMed IDs with MeSH annotations  for example: \url{http://eutils.ncbi.nlm.nih.gov/entrez/eutils/esearch.fcgi?db=pubmed&mindate=2014&maxdate=2015/11/19&term=+AND+\%28GENES\%5Bmesh\%5D+\%29}; and one that fetches the abstracts from the Ids retrieved from the previous URL, for example: \url{http://eutils.ncbi.nlm.nih.gov/entrez/eutils/efetch.fcgi?db=pubmed&retmode=xml&id=26580448,26580161,26575237,26577665}.

Finally, in the Answer Builder module, we use yet another Web Service where we use Linked Life Data and their SPARQL query endpoint, to generate rdf triples according to the ranked abstracts and annotations.\footnote{\url{http://linkedlifedata.com}})

%** FCOUTO: por de forma a que os exemplos se percebam no texto e não obriguem o leitor a ir ver a footnote. talvez divividir o urls em varias partes

%** FCOUTO: cuidado com o sudo se siglas deve ser sempre igual: PubMed Id vs. PubMed ID 

\begin{figure*}[ht!]
\label{sysarchitecture}
\centering
\includegraphics[width=\textwidth]{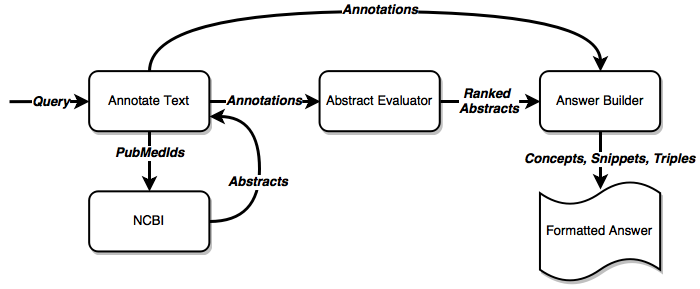}
\caption{The pipeline of WS4A with its main modules.}
\end{figure*}

\section{Pipeline} \label{pipeline}
Figure 1 shows the modular nature of WS4A, where each module was designed to be as much independent from the others as possible. 

% The main modules will be somewhat detailed in the following sections.\par
%\subsection{Saving Data} (tirar?)
%The first module that is present is the recorder.py (talvez não dizer nome dos ficheiros) ** sim retirar, where the data is to be stored, along with other important variables useful to many operations along the way. This first module runs after the file is read as a JSON file to memory. The flow of the data is already described in \ref{sysarchitecture}. \par
%\subsection{Query Annotation}

We have a module, Annotate Text, where question annotations by ontologies takes place. In this module, the Web Services used are the ones described in Section \ref{webservices}, in particular the BioPortal web services, which returns the annotations by their respective ontologies. Following the annotation retrieval, each annotation is used to  gather the respective PubMed IDs related to these annotations using the NCBI module, which also includes a local data structure containing the CheBi ontology. %\subsection{Filtering Abstracts}

%** FCOUTO: described in 2 !?

%** FCOUTO: não se percebe o que o modulo faz, é preciso dizer que web services usa e como? talvez a secção web services devai estar junto com este modulo

After the PubMed IDs are gathered into one structure, only the ten most recent articles are used, independently of the ontology it comes from. This measure was taken just to improve temporal performance. %\subsection{Abstract Annotation}
The Abstract Annotator module comes in use: we take these 10 abstracts in order to obtain annotations from the abstracts per ontology. This module is in fact the Annotate Text module, which reuses the same functions and methods present previously, but changing the input to abstracts instead of questions. %\subsection{Abstract Evaluation}

%** FCOUTO: um abstract não vem de uma ontologia -> abstracts per ontology

%** FCOUTO: This measure: qual measure? -> only the ten most recent articles are used

%** FCOUTO: não se percebe -> e agora?

In the Abstract Evaluator module, only the abstracts that are deemed to contain useful information are used in the  response. For each abstract, the following scores are registered for each abstract:
\begin{itemize}
\item Jaccard score between the query and abstract annotations \footnote{\url{http://ag.arizona.edu/classes/rnr555/lecnotes/10.html}}
\item hierarchical distance score between the query and abstract annotations
\item frequency between the top semantic annotations of the abstract and query
\item sentence semantic similarity score \footnote{\url{http://sujitpal.blogspot.pt/2014/12/semantic-similarity-for-short-sentences.html}}
\end{itemize}

%** FCOUTO: a primeira frase tb é super confusa

%** FCOUTO: porque aqui o link do arizona nao esta em footnote!?

After these steps, a grade is given to the abstract using these scores. Depending on the operation being computed (training or classification of abstracts), there are two possible courses of action: approval (or disapproval) of the abstract, so it can take part in the response in case of being in the training phase; otherwise, it adds the scores to a main structure, along with the abstract to be used to generate the n-grams. 

%** FCOUTO: dizer quais 3 valores 

%** FCOUTO: course repetido duas vezes

%** FCOUTO: tb nao se precebe as 3 categorias, dif com valores

After ranking and selecting the abstracts, we use another module, Answer Builder, that selects the 10 (according to the BioASQ's rules)  best valued snippets from the semantic analysis. 
Also, we generate the concepts required for the response. We take the initial annotated terms from the query with no regard of ontology. WS4A does the same to the annotations from the abstracts, but sums up the semantic similarity from each annotation, selecting the top 10 annotations by their score. %\subsection{RDFing}

%** FCOUTO: o according to Bioasq rules paarec indicar que exitem uma regras a usar, mas é só um limite

%** FCOUTO: não misturar we com WS4A, é melhor dizer sempre WS4A

The following step consists of taking these top 10 annotations from the abstracts, intercept them with the original annotations. With this result, we use only the MeSH annotations, to generate the RDF triples. To filter the resulting great amount of rdf triples, we use tf-idf in order to obtain those top 10 that add more content. 

The challenge provided training and test datasets that we used for our machine learning approach. %Those datasets consisted of gold answers, snippets, documents and RDF triples. 
In CLEF 2015 \cite{clef2015} among the systems that entered the BioASQ task 3b \cite{balikas2015results}, we noted that OAQA \cite{OAQA2015} used supervised learning techniques to predict answer types (answer type coercion) and score candidate answers.
We used Support vector machines \cite{joachims1999making,pedregosa2011scikit} to classify the relevance of the abstracts to a given query. From the Abstract Evaluation module  we select to work as features the four scores of the Abstract Evaluator module and the top 5000 n-grams.

\section{Results}
\begin{table*}
\begin{center}
\begin{tabular}{| c | c | c | c | c | c | c | c | c | c |}
\hline
\multicolumn{2}{| c |}{\multirow{2}{*}{\textbf{Batch}}} & \multicolumn{2}{| c |}{\textbf{Mean Precision}} & \multicolumn{2}{| c |}{\textbf{Mean Recall}}  & \multicolumn{2}{| c |}{\textbf{Mean FMeasure}} & \multicolumn{2}{| c |}{\textbf{Time (sec)}}   \\ 
\cline{3-10}
\multicolumn{2}{| c |}{} &  ML & w/o ML & ML & w/o ML & ML & w/o ML & ML & w/o ML \\
\hline
\multirow{2}{*}{1} & Docs & 0.010 & 0.010 & 0.012 & 0.016 & 0.010 & 0.012 & \multirow{2}{*}{16 125} & \multirow{2}{*}{8690} \\ \cline{2-8}
 & Snippets & 0.004 & 0.004 & 0.005 & 0.006 & 0.004 & 0.005 &  & \\ \hline
\multirow{2}{*}{2} & Docs & 0.005 & 0.003 & 0.002 & 0.012 & 0.003 & 0.005 & \multirow{2}{*}{11 130} & \multirow{2}{*}{8952} \\ \cline{2-8}
 & Snippets & 0.000 & 0.001 & 0.000 & 0.004 & 0.000 & 0.001 & &  \\ \hline
\multirow{2}{*}{3} & Docs & 0.009 & 0.005 & 0.006 & 0.004 & 0.007 & 0.004 & \multirow{2}{*}{10 005} & \multirow{2}{*}{8900} \\ \cline{2-8}
& Snippets & 0.002 & 0.000 & 0.002 & 0.000 & 0.002 & 0.000 &  &  \\ \hline
\multirow{2}{*}{4} & Docs  & 0.002 & 0.001 & 0.002 & 0.002 & 0.002 & 0.002 & \multirow{2}{*}{11 940} & \multirow{2}{*}{8967} \\ \cline{2-8}
& Snippets & 0.001 & 0.001 & 0.001 & 0.001 & 0.002 & 0.001 &  &  \\ \hline
\multirow{2}{*}{5} & Docs & 0.007 &  0.005  & 0.013 & 0.013 & 0.002 & 0.005 & \multirow{2}{*}{9790} & \multirow{2}{*}{8358} \\ \cline{2-8}
& Snippets & 0.001 & 0.003 & 0.001 & 0.004 & 0.001 & 0.003 &  &  \\ \hline
\end{tabular}
\end{center}
\label{xpto}
\caption{WS4A results for all batches using the final version}
\end{table*}

Table 1 shows WS4A results for every batch and using the final version of the system. The times were obtained in a desktop computer equipped with an Intel(R) Core(TM)2 Duo CPU and 6 GB RAM.

%We present the results with and without using Machine Learning to measure its impact. During the challenge, from iteration 1 to iteration 3 we submitted the results using a preliminary version of WS4A, that contained several implementation issues that were resolved at the time of iteration 5. In iteration 4, we have not submitted results. In iteration 5 WS4A was fully functional, and included also the application of SVM in phase A. The models used in batches 1 to 3 were the same, with machine learning classification used only in the last two. Unfortunately, this was not used in other aspects and more profoundly due to time limitations.

Our results, among all participants placed on the lower half of the result table, with the exception of the fourth batch results in which we did not submit any result set. The best result F-measure scored 0.24 in average, and so the overall results were also not very high. This results reflect phase A. In phase B, we were able to achieve a top 3 result for the first batch, obtaining lower level table results for the other batches. We achieved two second places in the first batch for the exact answers and ideal answers category.

Comparing to last year's results \cite{balikas2015results}, WS4A was far from achieving a comparable performance mainly because we registered for this competition some weeks before the date of the first batch release and no time for tuning the system was available. Also, in this year's competition, the type of answer wanted would already be given, so last year's results are not strictly comparable comparing to ours. 

The average time to answer a question was of around 90 seconds, with slow internet connection sometimes piking the time for obtaining an answer. W4SA was not developed to obtain highly accurate results but instead to use as much web services as possible. We demonstrated that such approach is feasible, however, we faced some issues related to availability faults in these services that hindered the progress of our system and the generation of the answers file.

We did not notice any other limitations, such as query limits (at least that we found or read about), but the servers would sometimes be unavailable for some reason for which we could never figure out the cause, and since this problem was quite random, we think it did not have anything to do with our research.

\section{Conclusions}
In this paper we showed the feasibility of developing a question and answering system mostly based on web services. Many of the techniques employed in WS4A have been used previously in other systems. For example, IIT \cite{IIT2015} retrieved documents using PubMed's web services like WS4A did, and extracted snippets based on similarity between sentences of the retrieved documents and the query. 
%HPI \cite{HPI2015} used term similarity between the terms of the question and the words of the documents, for snippet retrieval purposes. 
%SNUMedinfo \cite{SNUMedinfo2015} takes passages and ranks them based on the number of unique keywords they contain. ** like we did with ontology terms?
LIMSI-CNRS \cite{LIMSI-CNRS2015} used WordNet relations (namely, synonyms) when comparing between words from the query and the answer choice, and the given short text. WS4A also used WordNet, when comparing between the words in the query and the words in the abstracts. The main difference between WS4A and the aforementioned systems, is that our system is mainly based on web services and fully explores the semantics given by the ontologies. Thus WS4A is a light system that can be easily deployed, and which is continuously updated given the extensive use of web services.

In the future, we intend to explore semantic similarity measures \cite{couto2013next}, and build a cache database in order to save and revisit results.
Snippet generation can also be improved through semantic similarity, as CoMiC \cite{CoMic2015} used semantic similarity for the short text's segmentation in the Entrance Exams task 2015 (using the C99 algorithm).
We also plan on improving the annotation gathering functionality, and include other sources such as DBpedia\footnote{http://dbpedia.org} and YAGO\footnote{http://www.mpi-inf.mpg.de/departments/databases-and-information-systems/research/yago-naga/yago/} for obtaining such annotations. YodaQA \cite{YodaQA2015} resorts to DBPedia Spotlight, a service that automatically annotates DBPedia concepts from plain text.

\section*{Acknowledgments}
This work was supported by FCT through funding of LaSIGE Research Unit, ref. UID/CEC/00408/2013.

%\pagestyle{headings}
% BibTeX users please use
\bibliographystyle{splncs}
\bibliography{references}

\end{document}